\pdfoutput=1

\documentclass[11pt]{article}

\usepackage[review]{acl}

\usepackage{times}
\usepackage{latexsym}

\usepackage[T1]{fontenc}

\usepackage[utf8]{inputenc}

\usepackage{microtype}

\usepackage{inconsolata}

\usepackage{graphicx}
\usepackage{comment}
\usepackage{array}
\usepackage{amsmath}
\usepackage{amssymb}
\usepackage{soul}
\usepackage{multirow}
\usepackage{subcaption}
\usepackage[dvipsnames]{xcolor}
\usepackage{tcolorbox}

\title{Long-Context Long-Form Question Answering for Legal Domain}

\author{Anagha Kulkarni~~~ Parin Rajesh Jhaveri~~~ Prasha Shrestha~~~ Yu Tong Han \\ \textbf{Reza Amini}~~~  \textbf{Behrouz Madahian} \\
        J. P. Morgan Chase \\
        \normalsize{\texttt{\{anagha.p.kulkarni, parinrajesh.jhaveri, prasha.shrestha\}@jpmchase.com}}, \\ \normalsize{\texttt{zorina.han@jpmorgan.com}, \texttt{\{reza.amini, behrouz.madahian\}@jpmchase.com}} }
\newcommand{\prompt}[3]{%
  \begin{figure*}[ht]
    \centering
    \vspace{1em}
    \noindent\colorbox{gray!20}{%
      \parbox{0.98\textwidth}{
        \textbf{#1:} \\ 
        #2
      }
    }
    \vspace{1em}
    \caption{#3}
  \end{figure*}
}

\nolinenumbers

\begin{document}
\maketitle
\begin{abstract}
Legal documents have complex document layouts involving multiple nested sections, lengthy footnotes and further use specialized linguistic devices like intricate syntax and domain-specific vocabulary to ensure precision and authority. These inherent characteristics of legal documents make question answering challenging, and particularly so when the answer to the question spans several pages (i.e. requires long-context) and is required to be comprehensive (i.e. a long-form answer).
In this paper, we address the challenges of long-context question answering in context of long-form answers given the idiosyncrasies of legal documents. We propose a question answering system that can (a) deconstruct domain-specific vocabulary for better retrieval from source documents, (b) parse complex document layouts while isolating sections and footnotes and linking them appropriately, (c) generate comprehensive answers using precise domain-specific vocabulary. We also introduce a coverage metric that classifies the performance into recall-based coverage categories allowing human users to evaluate the recall with ease. We curate a QA dataset by leveraging the expertise of professionals from fields such as law and corporate tax. Through comprehensive experiments and ablation studies, we demonstrate the usability and merit of the proposed system.
\end{abstract}

\section{Introduction}

\label{sec: intro}

Legal documents are distinguished by their intricate structure, often comprising numerous nested sections and subsections as well as extensive footnotes and references to support the main text. The topic of question answering (QA) with legal documents has been studied before \cite{abdallah2023exploring, chakravarty2019improving, yang2024large}. Still several significant challenges remain unaddressed. For instance, if underlying layout elements such as section headers and footnotes in these documents are not accurately parsed, important contextual information may be lost. Further, the language used by experts in the legal domain is intentionally complex, utilizing sophisticated syntax as well as precise, domain-specific terminology. 

In this paper, we focus on addressing the challenges of producing long-form answers that demand an understanding across an extended context for legal documents. Long-form QA requires producing detailed, multi-sentence, paragraph-level responses to complex, open-ended questions. Additionally, long-context QA entails analyzing and synthesizing information from multiple lengthy passages to formulate a complete answer. 

\begin{figure*}[th!]
\centering
\includegraphics[width=0.9\textwidth,scale=0.99]{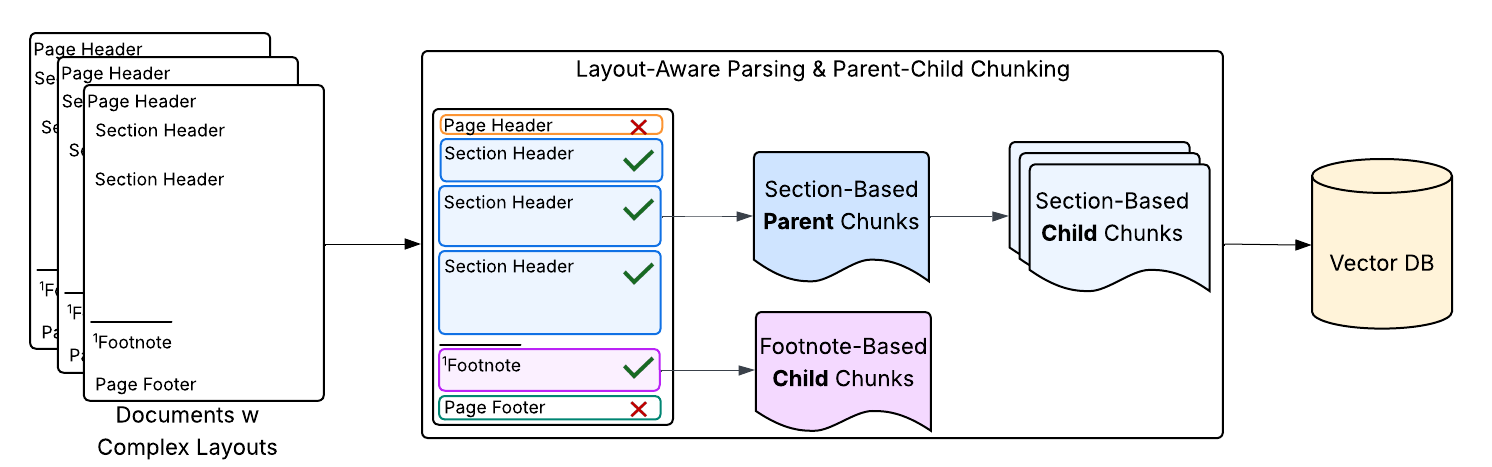}
\caption{
Overview of LCLF-QA ingestion: Layouts of legal documents are parsed into page headers, page footers, sections, and footnotes. Page headers and footers are filtered out, sections and footnotes are used to create parent and child chunks: (a) Each reasonably sized section becomes a parent chunk, and is divided into child chunks of appropriate lengths. (b) Footnotes on a page are grouped as a child chunk, and linked to parent chunks on that page.
}
\label{fig:lclfqa_1}
\end{figure*}

\begin{figure*}[th!]
\centering
\includegraphics[width=0.99\textwidth,scale=0.99]{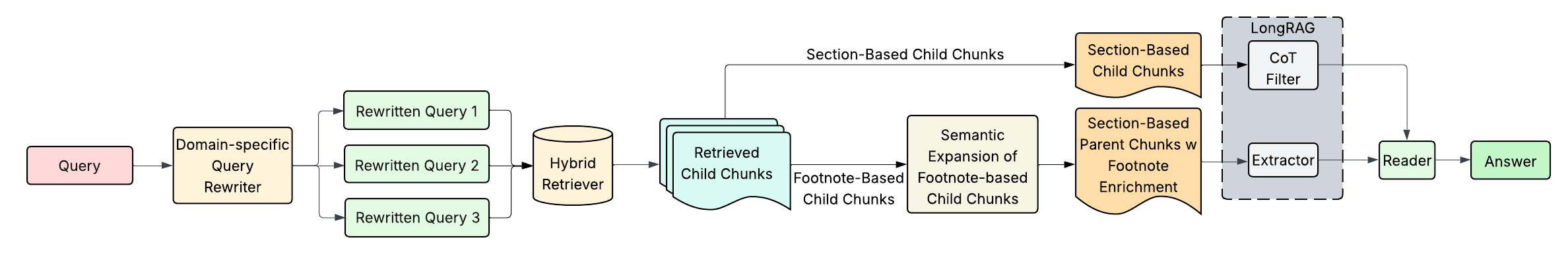}
\caption{
\textcolor{black}{Overview of LCLF-QA inference: Domain-specific query rewriter provides effective retrieval by reducing query ambiguity. The query and its rewrites retrieve relevant child chunks. During semantic expansion, retrieved footnote chunks are linked to parent chunks and parents are injected with footnote content. Retrieved child chunks are sent to CoT filter, parent chunks to extractor. Their outputs are used by domain-specific reader to form an answer.
}
}
\label{fig:lclfqa_2}
\end{figure*}

To understand the challenges of long-context long-form question answering, consider the following questions sourced from legal domain experts: (a) What is the available guidance regarding the US withholding tax treatment of <financial-product-name>, and at what level-of-comfort? (b) What are the potential tax characterizations of a <financial-product-name>? Given these examples, the challenges include:

\begin{itemize}
\item For question (a), the underlying intent of the question needs to be reinterpreted in the context of the existing sources. For example, the term ``level-of-comfort" which is not present in the source document needs to be reinterpreted to retrieve relevant information. This is especially important when users use specialized terminology without context, which can lead to misinterpretation of the question. Further, if the financial product name is an acronym, it will require expansion to capture appropriate context. 

\item For question (b) the system should be capable of extracting implicit context (such as latent facts from section headers and footnotes in the structural elements of the layout) as well as explicit context. This is crucial when there are similar-sounding facts that offer subtly different information. Therefore, reconciling the context with the details is quite crucial. 

\item Ultimately, the generated answer should follow the intricacies of the domain. For question (a), the answer must use precise vocabulary like ``should'' level-of-comfort, where ``should'' is a technical term used to reflect strong support for a tax opinion, instead of vague non-technical words like ``reasonable'' or ``high''. 
\end{itemize}

There are a few prior works which optimize for long-form QA in the legal domain \cite{abdallah2023exploring, nigam2023legal, louis2024interpretable, ujwal2024reasoning}. However, these ignore the problem of long-context as well as of latent context in document layouts. Therefore, none of the prior works focus on the aforementioned challenges associated with long-context long-form QA on real-world legal documents with complex and noisy document layouts.

To address these challenges we propose Long-Context Long-Form QA (LCLF-QA) illustrated in Figures \ref{fig:lclfqa_1} and \ref{fig:lclfqa_2}. LCLF-QA builds on top of \citet{zhao2024longrag}'s LongRAG architecture, which is the SOTA for long-context QA. However by itself, it fails to address the aforementioned challenges as shown in Table \ref{tab:main_results}. Our main contributions are: 

\begin{itemize}

\item \textbf{Domain-Specific Query Re-writer}~~ This component within LCLF-QA integrates domain-specific knowledge into user queries, facilitating more accurate retrieval. It is useful when queries contain special terms that are missing in the source text. 
\item \textbf{Layout-Aware Smart Chunking}~~ This component within LCLF-QA captures latent information from the structural layout elements like section headers, footnotes. The addition of this component helps enrich the retrieved chunks. 
\item \textbf{Recall-based Coverage Metric}~~ This metric allows for categorization of recall coverage into complete, partial and insufficient recall. This categorization helps when humans are evaluating recall of the generated answers. 

\end{itemize}



In the following sections, we outline the problem statement, provide a detailed explanation of the proposed method, report comprehensive evaluations and discuss relevant prior literature.

\section{Preliminaries}


\paragraph{RAG-Based QA.} Let $\mathcal{D}$ be a set of raw PDFs, where $\mathcal{D} = \{D_1, D_2, \dots, D_N \}$, $q \in \mathcal{Q}$ be a query, where $\mathcal{Q}$ is the universe of queries; $\mathcal{A}$ be the universe of answers to $q$. Let $C_i$ be a list of chunks obtained by segmenting the document $D_i$ into smaller pieces, similarly, we have $\mathcal{C}$ which is a collection of chunks corresponding to all documents in $\mathcal{D}$, such that, $|\mathcal{C}| >> |\mathcal{D}|$. Let $\mathbb{R}$ be a RAG-based \cite{lewis2020retrieval} QA system, with components retriever $\mathcal{R}$ and generator $\mathcal{G}$, we have, $\mathbb{R}(~\mathcal{G}~(q, ~\mathcal{R}~(q, ~\mathcal{C}))) \models a$, i.e. the retriever takes $q$ and $\mathcal{C}$ as input and produces a list of top-$k$ relevant chunks $\mathcal{C}_\mathcal{R}$, which is consumed by the generator along with $q$ to produce the answer $a$.

\paragraph{LongRAG-Based QA.} The objective in long-context QA is to capture long-context relationships. 
To that end, \citet{zhao2024longrag} propose a long-context extractor $\Sigma$ and a chain-of-thought (CoT) guided filter $\Phi$, designed to extract global information $I_g$ and identify factual details $I_d$, respectively. To enable these components, \citet{zhao2024longrag} organize the chunks into parent-child relationships. Let $\mathcal{C}_c$ be the collection of child chunks, while $\mathcal{C}_p$ be a collection of parent chunks, such that, $\forall c \in \mathcal{C}_c, \forall p \in \mathcal{C}_p, c \subset p \land |p_i| >> |c_i|$, that is, each child chunk is subsumed by its parent chunk and is smaller in size. Their retriever, $\mathcal{R}(q, \mathcal{C}_c) = \{\mathcal{C_R}_c \cup \mathcal{C_R}_p |~\forall c \in \mathcal{C_R}_c, ~\exists p \in \mathcal{C_R}_p : c \subset p\}$, returns retrieved child chunks, $\mathcal{C_R}_c$ and corresponding parent chunks, $\mathcal{C_R}_p$.
Extractor, $\Sigma$, extends the semantic memory using retrieved parent chunks $\mathcal{C_R}_p$ by organizing them into a long context.
Meanwhile filter, $\Phi$, performs a reasoning-based filtering on retrieved child chunks $\mathcal{C_R}_c$. This involves generating a CoT to check relevancy of child chunks $\mathcal{C_R}_c$, followed by filtering out irrelevant ones. The output of $\Sigma$ (i.e. long context) and $\Phi$ (relevant child chunks) is then forwarded to the generator, $\mathcal{G}$. \citet{zhao2024longrag}'s system $\mathbb{LC}$ gives, $\mathbb{LC}(~\mathcal{G}~(q, ~\Sigma~(q, \mathcal{C_R}_p), ~\Phi~(q, \mathcal{C_R}_c))) \models a$.

\section{LCLF-QA}

LCLF-QA tackles the challenges with real-world legal documents mentioned in Section \ref{sec: intro} by introducing three main components:

\begin{figure*}[th!]
    \centering
    \begin{subfigure}[t]{0.3\textwidth}
        \centering
        \includegraphics[width=0.99\textwidth]{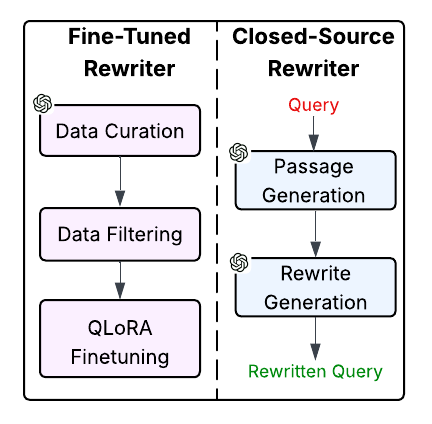}
        \caption{Left: Steps for fine-tuning an open-source model for query rewriting. Right: Steps for generating rewrites using a closed-source model.}
    \label{fig:dsqr_1}
    \end{subfigure}%
    ~ 
    \begin{subfigure}[t]{0.7\textwidth}
        \centering
        \includegraphics[width=0.99\textwidth]{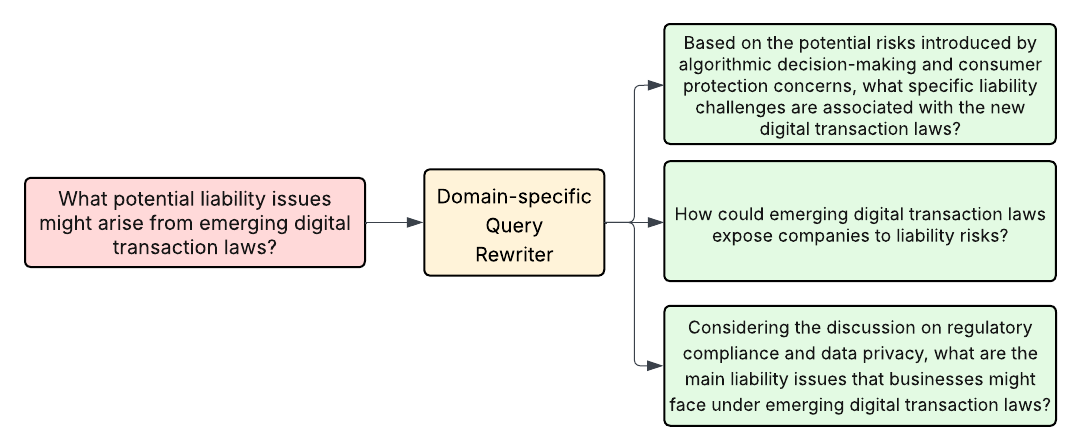}
        \caption{An example of query rewrites generated using our domain-specific query rewriter}
    \label{fig:dsqr_2}
    \end{subfigure}%
    \caption{Domain-Specific Query Rewriter}
\label{fig:dsqr}
\end{figure*}

\paragraph{Domain-specific query re-writer.} The effectiveness of retrieval in RAG depends on how well the query is constructed. User queries tend to be ambiguous and may lack sufficient context for accurate retrieval. Furthermore, the language in the queries may not always match the source documents, containing acronyms or shortened phrases. These kind of situations are particularly prevalent in domain-specific scenarios. To that end, a domain-specific query re-writer can be employed to transform a user query by (a) expanding acronyms and short forms, (b) adding broader context, and (c) paraphrasing the intent of the user query to provide diversified perspective. 
Refer to Figure \ref{fig:dsqr_2} for an example of rewritten query.

\paragraph{Layout-aware smart chunking.} Legal documents contain dense information distributed across complex layouts and they rely heavily on structural cues, such as section headers and footnotes. Section headers serve as topical boundaries and provide critical context for interpreting the section content. Footnotes in legal documents provide essential clarifications, definitions, and exceptions that support the main content. Standard chunking strategies based on token count or sentence boundaries struggle to preserve the structural information leading to semantically incoherent chunks that compromise the retrieval. They also fail to capture the relationship between the content and footnotes. To that end, we devise a chunking strategy that segments the document based on sections, similar to \citet{yepes2024financial}, but adapted to parent-child chunking. Further, we enhance the chunks with section headers and footnotes.

\paragraph{Domain-specific generator.} In legal domain, it is crucial to respond to the user in precise vocabulary for the sake of precision. The domain-specific generator is parameterized with domain-specific vocabulary as well as examples of answer-styles preferred by users. In addition, the generator uses the output of the extractor and the CoT guided filter along with original question, rewritten version of the questions as well as few shot examples of domain-specific QA to output the answer.

Formally, domain-specific query re-writer $\zeta_t$, takes a query and produces rewrites $\zeta_t(q) = \{ \hat{q_1}, \hat{q_1}, \dots, \hat{q_l}\}$, where $t$ is the domain-specific context that the query re-writer is parameterized with. Let $q_\zeta = \{q \cup \zeta_t(q)\}$ be a union of these queries. Here, $t$ can stand for fine-tuning of an open-source model or domain-specific prompt context of a closed-source model as shown in Figure \ref{fig:dsqr_1}. Layout-aware smart chunking uses semantically enriched section-based and footnote-based parent and child chunks as shown in Figure \ref{fig:lclfqa_1}. Let $\Delta_p$ and $\Delta_c$ be the collection of such enriched child and parent chunks. The hybrid retriever, $\mathcal{R}_h(q_\zeta, \Delta_c) = \{\Delta_{\mathcal{R}_p} \cup \Delta_{\mathcal{R}_c} \}$, retrieves these child chunks and parent chunks by leveraging semantic and lexical relationships. The domain-specific generator $\mathcal{G}_\mu$ is a generator parametrized by domain-specific knowledge and/or examples of user-specific answer styles $\mu$. Together we have, LCLF-QA$(~\mathcal{G}_\mu~(q_\zeta, ~\Sigma(q_\zeta, \Delta_{\mathcal{R}_p}), ~\Phi(q_\zeta, \Delta_{\mathcal{R}_c}))) \models a_\ell$, where $a_\ell$ is a long-form answer to $q$.

 

\subsection{Domain-Specific Query Re-writer}

We discuss two approaches for query re-writer: fine-tuning-based and prompting-based.

\subsubsection{Fine-Tuned Query Re-writer}
\label{sec: fine-tuned query rewriter}
We fine-tuned a Mistral-3B-Instruct model using QLoRA \cite{dettmers2023qlora} with LoRA parameters: $r=64$ and $\alpha = 128$, for our re-writer component. During training, to ensure that there is less noise, we mask the instructions when calculating loss and use standard cross entropy loss.
We devise a process of data curation and filtering to enhance the quality of the dataset used to fine-tune the open source query re-writer. 
Below, we delve into the specific steps of data curation and filtration.


\paragraph{Data Curation.} For fine tuning, we require $\langle q, \hat{q}\rangle$ pairs in the order of thousands of examples. Due to lack of availability of human-annotated data, we generate synthetic query dataset as discussed in Appendix \ref{sub:synthetic_generation}. These queries along with the source chunks are rewritten into unambiguous re-writes using GPT-4o \cite{hurst2024gpt}. The prompt for rewrite generation is included in Appendix \ref{sec: prompts}.

\paragraph{Data Filtering.} Our analysis shows that the effectiveness of a query rewrite is closely tied to the type of retriever, and there is no universal rewrite that works optimally across all systems. This challenge is particularly pronounced in domain-specific tasks, where dense retrievers often lack specialization. To address this, we filter $\langle q, \hat{q} \rangle$ pairs by comparing the rank of the source document $D$ for both $q$ and $\hat{q}$, retaining only those pairs where $\hat{q}$ improves $D$'s rank in both sparse and dense retrieval: $\text{rank}^{\hat{q}}_{\text{sparse}}(D) > \text{rank}^{q}_{\text{sparse}}(D)$ and $\text{rank}^{\hat{q}}_{\text{dense}}(D) > \text{rank}^{q}_{\text{dense}}(D)$. This ensures that the rewrites are well-matched to the retriever’s capabilities, resulting in higher quality data for QLoRA-based fine-tuning.



\subsubsection{Closed-Source Query Re-writer}
To leverage the extensive internal knowledge embedded within closed-source LLMs, we developed a query re-writer using GPT-4o. Our approach follows \citet{wang-etal-2023-query2doc}'s Query2doc framework, which involves a two-step process to enhance query generation. The LLM is first prompted to generate a document that answers the user query, allowing it to extract internal knowledge that it believes is relevant to the query. We augment the prompt given in Query2doc to contain domain-specific few-shot examples to help with in-context learning. This allows the LLM to learn the domain-specific nuances of a user query and generate a relevant document within the same domain. Subsequently, the generated document is fed back into the LLM to produce an expanded and more detailed query. This refined query benefits from the domain-specific context provided by the document, ensuring that it is not only comprehensive but also tailored to the specific nuances of the subject matter. 

\subsection{Layout-Aware Smart Chunking}


\paragraph{Section-Based Parent Chunks.} Parsed document, $\mathcal{P}(D_i)$, is segmented into a list of sections $S_i$, where $S_i = \{s_0, \dots, s_m\}$. A parent chunk, $\delta_p$ is a set of sections, $\delta_p = \{s_a \cup \ldots \cup s_b\}$, such that, $|s_a \cup \ldots \cup s_{b-1}| < L$ and $|s_a \cup \ldots \cup s_{b}| \geq L$, i.e., sections are merged until they exceed the maximum parent chunk size, $L$.
A section overlaps when the size of the last section in $\delta_p$ is less than $L$, i.e., if $s_a$ is the last section in ${\delta_p}_{i-1}$, and $s_{a} < L$ then ${\delta_p}_{i}$ starts with $s_{a}$ else with $s_{a+1}$.
This overlapping strategy maintains semantic continuity across chunks. To ensure logical continuity (a) the page headers and footers are filtered out, and (b) the footnotes are stored as parent chunk metadata to be used during the child chunking process.

\paragraph{Section-Based and Footnote-Based Child Chunks.} A parent chunk $\delta_p$ is recursively segmented into smaller child chunks while maintaining sentence boundaries. To facilitate effective retrieval of a child chunk, the text within each child chunk is mapped back to $\delta_p$ to extract the corresponding section headers. These section headers are then injected into the text of the child chunk using tags, such as \texttt{<section-header>}. These enriched segments are referred to as section-based child chunks, $\delta_c^s$. Additionally, the footnotes associated with the parent chunk $\delta_p$ are separately mapped to form footnote-based child chunks, $\delta_c^f$. The collection of child chunks, $\Delta_c$, consists of both section-based and footnote-based child chunks.

\begin{figure}[ht!]
\centering
\includegraphics[width=0.49\textwidth]{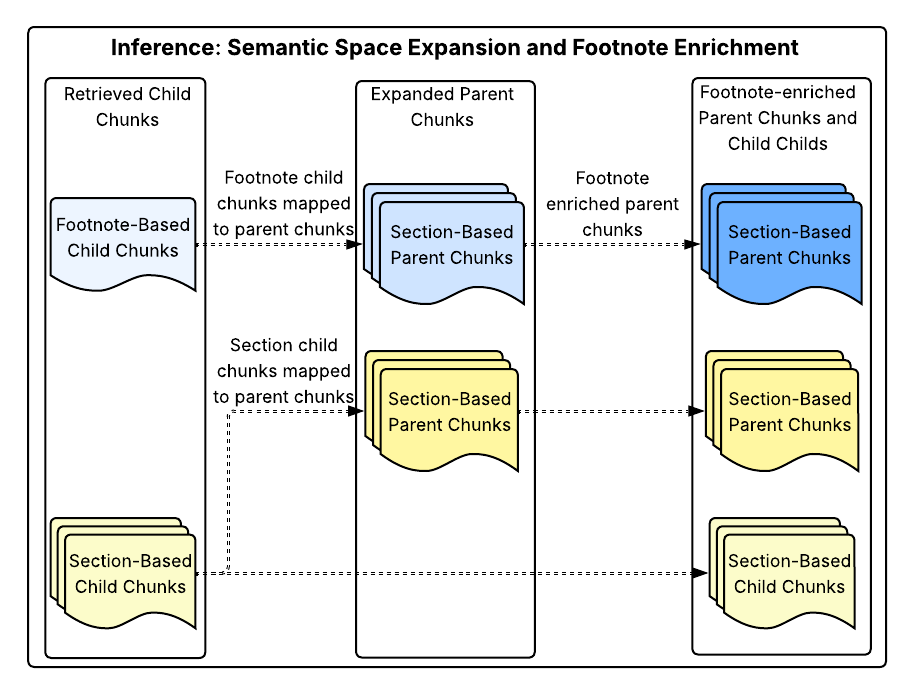}
\caption{
Retrieved footnote-based child chunks are used to retrieve the linked section-based parent chunks. These section-based parent chunks go through footnote enrichment, where the footnote is appended using tags.
}
\label{fig:chunking}
\end{figure}

\paragraph{Footnote-based enrichment.} During retrieval, if a footnote-based child chunk, $\delta_c^f$, is retrieved, then the semantic space of retrieved chunks is expanded to encompass all section-based parent chunks that are associated with $\delta_c^f$ as shown in Figure \ref{fig:chunking}.  
Further, to provide a complete context to extractor, $\Sigma$, the footnote is injected into associated $\delta_p$ using tags like \texttt{<footnote>}.

\subsection{Hybrid Retrieval}

\label{sec: hybrid retrieval}
To identify semantically and syntactically relevant content for long-form answer generation, we employ a hybrid retrieval approach that combines both sparse and dense retrieval methods.
This combination enables us to leverage the precision of exact lexical matches and the semantic generalization capabilities of neural embeddings.
We use BM25 \cite{amati2018bm25} as our sparse retriever, which scores the child chunks, \(\Delta^c\), based on exact term overlap and inverse document frequency and OpenAI's `text-embedding-3-large' model as our dense retriever. This model produces a 3072-dimensional embedding for each chunk and computes similarity using cosine distance. The chunk corpus is pre-embedded and indexed using FAISS \cite{douze2024faiss} for efficient dense nearest-neighbor search.
Let $\mathcal{R}_{sparse}(q_\zeta, \Delta_c) = \{\Delta_{{sparse}_p} \cup \Delta_{{sparse}_c} \}$ be the set of retrieved chunks for sparse retriever and equivalently for the dense retriever, we have, $\mathcal{R}_{dense}(q_\zeta, \Delta_c) = \{\Delta_{{dense}_p} \cup \Delta_{{dense}_c} \}$.

To combine these results, we use Reciprocal Rank Fusion (RRF) \cite{cormack2009reciprocal} which computes the fusion score for a chunk $\delta_c \in \Delta_c$  as:
\[
\mathbf{RRF}(\delta_c) = \sum_{\mathcal{R} \in \{\mathcal{R}_{sparse}, \mathcal{R}_{dense}\}} \frac{1}{\lambda + {rank}_{\mathcal{R}}(\delta_c)}
\]
where $rank_\mathcal{R}(\delta_c)$ is the rank of $\delta_c$ in retrieval list of $\mathcal{R}$, and $\lambda$ is a smoothing constant (typically $\lambda = 60$). We retrieve the top-$k$ child chunks $\{{\delta_c}_1, {\delta_c}_2, \dots, {\delta_c}_k\}$ based on the RRF score.

\section{Experiments}

This section reports the comparison of our main results against baselines, ablations of individual components and qualitative analysis. 

\paragraph{Dataset}
\label{sec: dataset}
Our dataset consists of 546 QA pairs in total. Among these, we curated a set of 60 pairs with the help of SMEs from the Legal and Corporate Tax sector. Each question maps to a single document and the questions span across 24 documents. 
These SME provided QA pairs are augmented with synthetically generated QA pairs that follow 6 different question formats as described in Appendix \ref{sub:synthetic_generation}. There are 486 synthetic QA pairs generated from 10 different documents. These pairs have been validated and corrected by human users to ensure high quality. 

\paragraph{Metrics.}

The SME gold answers are terse in nature (and the synthetic answers are styled similarly), so we only check recall when comparing our answers against the gold answers for a fair comparison. For computing the recall, we modified RAGChecker \cite{ru2024ragchecker} prompts to extract nested triples which not only capture atomic facts but also preserve implications. We augment the prompt with domain specific examples. Recall is computed as the number of claims entailing correctly over the total number claims in the gold answer.

In addition to recall, we introduce a coverage metric that checks if the information in the gold answer is present in the generated answer. This involves (1) extracting the set of claims from the gold answer that are necessary to answer the question, (2) determining if these claims are present in the generated answer. Based on CoT reasoning, the generated answer is categorized into: (a) Complete: all necessary claims are present in the generated answer, (b) Partial: some necessary claims are missing from the generated answer, (c) Incorrect: some necessary claim is incorrect in the generated answer. The prompt for assigning a category is in Appendix \ref{sec: prompts}. 
We use these categories to assign a score to each question answer pair: Complete $\rightarrow$ 2, Partial $\rightarrow$ 1, Incorrect $\rightarrow$ 0. A maximum possible score is achieved when all answers deemed ``Complete". Coverage score is calculated by taking a sum of the scores of all QA pairs normalized by the maximum possible score,  $\text{Coverage Score} = \frac{2 \cdot N_{\text{Complete}} + N_{\text{Partial}}}{2 \cdot N_{\text{Total}}}$ 
where $N_{\text{Complete}}$ is the number of Complete answers, $N_{\text{Partial}}$ is the number of Partial answers, and $N_{\text{Total}}$ is the total number of QA pairs.

\subsection{Main Results}

\begin{table}[t]
    \centering
    \resizebox{\columnwidth}{!}{%
    \def\arraystretch{1.2}
    \begin{tabular}{|l|l|r|r|r|l|}
        \hline
        & Recall & Complete & Partial & Incorrect & Score  \\ 
        \hline
        RAG & 0.5369 & 96 & 427 & 23 & 0.5668 \\
        \hline
        LongRAG & 0.6486 & 182 & 340 & 24 & 0.6446 \\
        \hline
        Ours: FT, q=1 & 0.6694* & 194 & 339 & 13 & 0.6658 \\
        \hline
        Ours: FT, q=3 & 0.6677 & 205 & 319 & 22 & 0.6676* \\
        \hline
        Ours: CS, q=1 & \textbf{0.6798**} & 200 & 334 & 12 & \textbf{0.6722*} \\
        \hline
        Ours: CS, q=3 & 0.6728* & 193 & 338 & 15 & 0.6630 \\
        \hline
    \end{tabular}
    }
    \caption{Comparison of LCLF-QA with baselines of RAG and LongRAG. FT and CS represent fine-tuned and closed source query re-writers with single or three rewrites. * Statistically significant compared to LongRAG at p$<$0.05; ** at p$<$0.01
}
    \label{tab:main_results}
\end{table}

Table \ref{tab:main_results}, shows the results of different LCLF-QA configurations as well as the baselines RAG and LongRAG. We report the results for: fine-tuned query re-writer and closed-source query re-writer both with single and multiple (three) query rewrites. The results have been averaged over three runs per QA pair in the dataset to avoid any model bias. Using the single query rewrite from the closed source re-writer, there is a maximum gain in recall over LongRAG, from 0.6486 to 0.6798. The results from the fine tuned query re-writer also shows good improvement over LongRAG with a recall of 0.6694 with the single query rewrite. The coverage metric also shows similar increase from LongRAG baseline to our pipelines, and follows the trend of lowering the total number of partial and incorrect answers and raising the number of complete answers across both datasets. 

\subsection{Ablation Studies}

As part of ablation studies, our objective is to test the following hypotheses: 

\noindent \textbf{H1:} \textit{LCQA primed with few-shot domain-specific examples performs poorly on domain-specific long-form QA compared to LCLF-QA.} 

\noindent \textbf{H2a:} \textit{LCLF-QA without a domain-specific re-writer exhibits poor answer quality.} 

\noindent \textbf{H2b:} \textit{LCLF-QA with multiple query rewrites is better than that with single query rewrite.} 

\noindent \textbf{H3:} \textit{LCLF-QA without layout-aware smart chunking exhibits poor answer quality.}  

\noindent \textbf{H4:} \textit{LCLF-QA without domain-specific parameterization of generator exhibits poor answer quality.} \\

\begin{table}[!t]
    \centering
     {\footnotesize
    \resizebox{\columnwidth}{!}{%
    \def\arraystretch{1.2}
    \begin{tabular}{|l|r|r|r|r|r|}
        \hline
        & Recall & Complete & Partial & Incorrect & Score  \\ 
        \hline
        H1  & 0.6312 & 136 & 379 & 31 & 0.5962 \\
        \hline
        H2a & 0.6610 & 195 & 327 & 24 & 0.6566 \\
        \hline
        H3  &   0.6662     & 193 & 339 & 14 & 0.6639 \\
        \hline
        H4  & 0.6556 & 175 & 358 & 13 & 0.6484 \\
        \hline
    \end{tabular}
    }
    }
    \caption{Ablation of pipeline components. C, P, I, score stand for complete, partial, incorrect, coverage score.}
    \label{tab:ablation}
\end{table}

For \textit{H1}, we wanted to see if LongRAG equipped with hybrid retrieval and few-shot example generator fails to solve the problem. Typically, priming the generator with few-shot examples can give good results. However, from the results of \textit{H1} in Table \ref{tab:ablation}, both recall and coverage score for the enhanced LongRAG are lower than those for our pipeline in Table \ref{tab:main_results}. We can see that the legal domain requires a specialized solution like LCLF-QA. For \textit{H2a}, we show results without a domain-specific re-writer, and again compared to the best pipeline in Table \ref{tab:main_results}, the results are lower for both recall and coverage score, thus proving the hypothesis that without re-writer the results will be poor. For \textit{H2b}, there is no conclusive result, as sometimes multiple query rewrites perform better than single rewrite and vice versa. This conclusion is supported by the results in Table \ref{tab:main_results}. For \textit{H3}, we show the results for LCLF-QA without layout-aware smart chunking, and we can see that this component is also crucial in improving the recall and coverage score. For \textit{H4}, we show results for LCLF-QA with a basic generator and we can see that this also makes a big difference in the quality of the results. Query re-writer, layout-aware chunking  and parameterized generator all individually add value to LCLF-QA.

\section{Related Work}

There are a few prior works which optimize for long-form QA in the legal domain \cite{abdallah2023exploring, nigam2023legal, louis2024interpretable, ujwal2024reasoning} but ignore the problem of long-context and of latent context in document layouts. 
Therefore, none of these works focus on challenges associated with long-context long-form QA on real-world legal documents with complex and noisy document layouts. 
The proposed LCLF-QA system derives value from two main components: domain-specific query re-writer and layout-aware chunking. 

\paragraph{Query Re-writer} In RAG systems, user queries play a vital role in information retrieval. However, users may not always provide enough details or clarity in their queries, leading to misunderstandings or misinterpretations by the system. In such cases, a query re-writer component can add value \cite{li2024dmqr, mao2024rafe, chan2024rq, wang2025maferw}. We build a fine-tuned open source query re-writer as well as a closed source query re-writer. For the closed source query re-writer, the prompt is parameterized with domain-specific knowledge, and for a given user query, an answer is generated. Subsequently, a re-write is produced based on this answer. \citet{wang2023query2doc}'s approach proposes the notion of $query \rightarrow doc  \rightarrow rewrite$. We borrow this idea and extend it to the legal domain.

\paragraph{Layout-based Chunking} Legal documents have a structured layout that contains a wealth of latent information. In a layout-agnostic chunking approach, this latent context gets lost. This can impact the overall quality of the chunks.
There are several works in literature \cite{yepes2024financial, tripathi2025vision, kiss2025max} that tackle the layout detection problem. We fine-tune a layout detection model which identifies page header, page footers, section header, footnotes, and text paragraphs and enrich the chunks using these elements.

To tackle the problem of long-context, we leverage \citet{zhao2024longrag}'s LongRAG architecture, which significantly outperforms several long-context LLMs and RAG-based approaches. We borrow the notion of global extractor which maintains coherent context across extended passages and that of the CoT guided filter which filters irrelevant chunks using CoT reasoning.

\section{Conclusion}
This paper presents a long-context long-form question answering system that addresses the challenges associated with legal domain. Our specialized query re-writing, layout-aware chunking, and generator parametrization strategies work together to improve recall and coverage score on a curated dataset. We see statistically significant gains against vanilla RAG and LongRAG. LCLF-QA is able to effectively reason across long contexts while working with the subtleties of domain-specific terms and complex document layouts.

\section*{Limitations}
Although our QA system is domain agnostic, currently, we have only evaluated it on legal documents in the corporate tax domain. We expect our results to translate well in other domains and have plans to verify them in the future. Except for one of our query re-writer components, all other components have only been tested with close source models and we acknowledge that this limits its application.

\section*{Acknowledgments}

We would like to thank Eshani Agrawal for her contributions to the layout-aware parsing codebase.

\bibliography{custom}

@article{yang2024large,
  title={Large language models for automated q\&a involving legal documents: a survey on algorithms, frameworks and applications},
  author={Yang, Xiaoxian and Wang, Zhifeng and Wang, Qi and Wei, Ke and Zhang, Kaiqi and Shi, Jiangang},
  journal={International Journal of Web Information Systems},
  volume={20},
  number={4},
  pages={413--435},
  year={2024},
  publisher={Emerald Publishing Limited}
}

@incollection{chakravarty2019improving,
  title={Improving the processing of question answer based legal documents},
  author={Chakravarty, Saurabh and Mehrotra, Maanav and Chava, Raja Venkata Satya Phanindra and Liu, Han and Krivansky, Matthew and Fox, Edward A},
  booktitle={Legal Knowledge and Information Systems},
  pages={13--22},
  year={2019},
  publisher={IOS Press}
}

@article{kiss2025max,
  title={Max--Min semantic chunking of documents for RAG application},
  author={Kiss, Csaba and Nagy, Marcell and Szil{\'a}gyi, P{\'e}ter},
  journal={Discover Computing},
  volume={28},
  number={1},
  pages={1--15},
  year={2025},
  publisher={Springer}
}

@article{tripathi2025vision,
  title={Vision-Guided Chunking Is All You Need: Enhancing RAG with Multimodal Document Understanding},
  author={Tripathi, Vishesh and Odapally, Tanmay and Das, Indraneel and Allu, Uday and Ahmed, Biddwan},
  journal={arXiv preprint arXiv:2506.16035},
  year={2025}
}

@article{yepes2024financial,
  title={Financial report chunking for effective retrieval augmented generation},
  author={Yepes, Antonio Jimeno and You, Yao and Milczek, Jan and Laverde, Sebastian and Li, Renyu},
  journal={arXiv preprint arXiv:2402.05131},
  year={2024}
}

@article{wang2023query2doc,
  title={Query2doc: Query expansion with large language models},
  author={Wang, Liang and Yang, Nan and Wei, Furu},
  journal={arXiv preprint arXiv:2303.07678},
  year={2023}
}

@inproceedings{wang2025maferw,
  title={MaFeRw: Query rewriting with multi-aspect feedbacks for retrieval-augmented large language models},
  author={Wang, Yujing and Zhang, Hainan and Pang, Liang and Guo, Binghui and Zheng, Hongwei and Zheng, Zhiming},
  booktitle={Proceedings of the AAAI Conference on Artificial Intelligence},
  volume={39},
  pages={25434--25442},
  year={2025}
}

@article{chan2024rq,
  title={Rq-rag: Learning to refine queries for retrieval augmented generation},
  author={Chan, Chi-Min and Xu, Chunpu and Yuan, Ruibin and Luo, Hongyin and Xue, Wei and Guo, Yike and Fu, Jie},
  journal={arXiv preprint arXiv:2404.00610},
  year={2024}
}

@article{mao2024rafe,
  title={Rafe: ranking feedback improves query rewriting for rag},
  author={Mao, Shengyu and Jiang, Yong and Chen, Boli and Li, Xiao and Wang, Peng and Wang, Xinyu and Xie, Pengjun and Huang, Fei and Chen, Huajun and Zhang, Ningyu},
  journal={arXiv preprint arXiv:2405.14431},
  year={2024}
}

@article{li2024dmqr,
  title={DMQR-RAG: Diverse Multi-Query Rewriting for RAG},
  author={Li, Zhicong and Wang, Jiahao and Jiang, Zhishu and Mao, Hangyu and Chen, Zhongxia and Du, Jiazhen and Zhang, Yuanxing and Zhang, Fuzheng and Zhang, Di and Liu, Yong},
  journal={arXiv preprint arXiv:2411.13154},
  year={2024}
}

@article{nigam2023legal,
  title={Legal question-answering in the indian context: Efficacy, challenges, and potential of modern ai models},
  author={Nigam, Shubham Kumar and Mishra, Shubham Kumar and Mishra, Ayush Kumar and Shallum, Noel and Bhattacharya, Arnab},
  journal={arXiv preprint arXiv:2309.14735},
  year={2023}
}

@inproceedings{ujwal2024reasoning,
  title={" Reasoning before Responding": Towards Legal Long-form Question Answering with Interpretability},
  author={Ujwal, Utkarsh and Surampudi, Sai Sri Harsha and Mitra, Sayantan and Saha, Tulika},
  booktitle={Proceedings of the 33rd ACM International Conference on Information and Knowledge Management},
  pages={4922--4930},
  year={2024}
}

@article{abdallah2023exploring,
  title={Exploring the state of the art in legal QA systems},
  author={Abdallah, Abdelrahman and Piryani, Bhawna and Jatowt, Adam},
  journal={Journal of Big Data},
  volume={10},
  number={1},
  pages={127},
  year={2023},
  publisher={Springer}
}

@inproceedings{louis2024interpretable,
  title={Interpretable long-form legal question answering with retrieval-augmented large language models},
  author={Louis, Antoine and van Dijck, Gijs and Spanakis, Gerasimos},
  booktitle={Proceedings of the AAAI Conference on Artificial Intelligence},
  volume={38},
  pages={22266--22275},
  year={2024}
}

@article{ru2024ragchecker,
  title={Ragchecker: A fine-grained framework for diagnosing retrieval-augmented generation},
  author={Ru, Dongyu and Qiu, Lin and Hu, Xiangkun and Zhang, Tianhang and Shi, Peng and Chang, Shuaichen and Jiayang, Cheng and Wang, Cunxiang and Sun, Shichao and Li, Huanyu and others},
  journal={Advances in Neural Information Processing Systems},
  volume={37},
  pages={21999--22027},
  year={2024}
}

@article{douze2024faiss,
  title={The faiss library},
  author={Douze, Matthijs and Guzhva, Alexandr and Deng, Chengqi and Johnson, Jeff and Szilvasy, Gergely and Mazar{\'e}, Pierre-Emmanuel and Lomeli, Maria and Hosseini, Lucas and J{\'e}gou, Herv{\'e}},
  journal={arXiv preprint arXiv:2401.08281},
  year={2024}
}

@incollection{amati2018bm25,
  title={BM25},
  author={Amati, Giambattista},
  booktitle={Encyclopedia of database systems},
  pages={323--326},
  year={2018},
  publisher={Springer}
}

@inproceedings{cormack2009reciprocal,
  title={Reciprocal rank fusion outperforms condorcet and individual rank learning methods},
  author={Cormack, Gordon V and Clarke, Charles LA and Buettcher, Stefan},
  booktitle={Proceedings of the 32nd international ACM SIGIR conference on Research and development in information retrieval},
  pages={758--759},
  year={2009}
}

@article{hurst2024gpt,
  title={Gpt-4o system card},
  author={Hurst, Aaron and Lerer, Adam and Goucher, Adam P and Perelman, Adam and Ramesh, Aditya and Clark, Aidan and Ostrow, AJ and Welihinda, Akila and Hayes, Alan and Radford, Alec and others},
  journal={arXiv preprint arXiv:2410.21276},
  year={2024}
}

@article{dettmers2023qlora,
  title={Qlora: Efficient finetuning of quantized llms},
  author={Dettmers, Tim and Pagnoni, Artidoro and Holtzman, Ari and Zettlemoyer, Luke},
  journal={Advances in neural information processing systems},
  volume={36},
  pages={10088--10115},
  year={2023}
}

@inproceedings{bolotova2022non,
  title={A non-factoid question-answering taxonomy},
  author={Bolotova, Valeriia and Blinov, Vladislav and Scholer, Falk and Croft, W Bruce and Sanderson, Mark},
  booktitle={Proceedings of the 45th International ACM SIGIR Conference on Research and Development in Information Retrieval},
  pages={1196--1207},
  year={2022}
}

@article{lewis2020retrieval,
  title={Retrieval-augmented generation for knowledge-intensive nlp tasks},
  author={Lewis, Patrick and Perez, Ethan and Piktus, Aleksandra and Petroni, Fabio and Karpukhin, Vladimir and Goyal, Naman and K{\"u}ttler, Heinrich and Lewis, Mike and Yih, Wen-tau and Rockt{\"a}schel, Tim and others},
  journal={Advances in neural information processing systems},
  volume={33},
  pages={9459--9474},
  year={2020}
}

@article{zhao2024longrag,
  title={Longrag: A dual-perspective retrieval-augmented generation paradigm for long-context question answering},
  author={Zhao, Qingfei and Wang, Ruobing and Cen, Yukuo and Zha, Daren and Tan, Shicheng and Dong, Yuxiao and Tang, Jie},
  journal={arXiv preprint arXiv:2410.18050},
  year={2024}
}

@inproceedings{wang-etal-2023-query2doc,
    title = "Query2doc: Query Expansion with Large Language Models",
    author = "Wang, Liang  and
      Yang, Nan  and
      Wei, Furu",
    booktitle = "Proceedings of the 2023 Conference on Empirical Methods in Natural Language Processing",
    year = "2023",
    url = "https://aclanthology.org/2023.emnlp-main.585/",
    pages = "9414--9423"
}

\appendix

\section{Appendix: Synthetic Query and QA Pair Generator}

\label{sub:synthetic_generation}

We synthesize domain-specific queries and QA pairs. The synthesized queries are used to fine-tune a query re-writer model described in Section~\ref{sec: fine-tuned query rewriter}, whereas the QA pairs are used to augment our evaluation dataset as discussed in Section~\ref{sec: dataset}. 

Given the task of either query or QA pair generation, this generator takes documents as input and performs following steps: (1) generate page-wise summary of the document (2) cluster the page-summaries into $k$ clusters using k-means clustering (3) pick a cluster at random and choose $n$ page-based chunks -- if $n$ chunks are not available in the cluster, pick another cluster at random until one with $n$ chunks is found (4) Use $n$ chunks to generate a query (or a QA pair) using one of following 6 types of question-styles:
(a) instruction-based: focuses on understanding the procedure/method of doing/achieving something, (b) reason-based: focuses on finding out reasons of/for something, (c) evidence-based: focuses on learning the features/description/definition of a concept/idea/object/event, (d) comparison-based: focuses on comparing two or more things, understanding their differences/similarities, (e) list-based: focuses on finding requirements of some process and (f) domain-specific: focuses on simulating question styles used by SMEs. We leverage the definitions and question templates provided by \citet{bolotova2022non} for question styles (a)-(d). This process runs for predetermined budget, $b$. 

\section{Appendix: Detailed Prompts}
\label{sec: prompts}

In the following we present the various prompts that were used in the LCLF-QA approach.

\prompt{Long Context Extractor Prompt}{
\# Instruction\\
You are an expert in corporate tax policies. \\
\\
\#\# Reference \\
<reference> \{context\} </reference> \\
\\
\#\# Rules\\
- Based on the above reference, please output the original information that needs to be cited to answer the question.\\
- Please ensure that the original information is detailed and comprehensive. \\
- The reference is delimited by <reference></reference>.  \\
- The question is delimited by <question></question>.\\ 
- A translation of the question in layman terms is delimited by <translated\_question></translated\_question>\\
- Your answer should be delimited by <information></information>.\\
\\
\#\# Question\\
<question> \{question\} </question>\\
<translated\_question> \{translated\_question\} </translated\_question>\\
\\
\#\# Information\\
}{Prompt for the long context extractor}

\prompt{Filter - CoT Prompt}{
\# Instruction\\
\\
\#\# Reference\\
\{context\}\\

\#\# Rules\\
- There are multiple reference texts.\\
- Your task is to give your thought process for the given question based on all the reference texts.\\
- Only output the thought process based on the reference texts.\\
- Each reference text is delimited by <reference></reference>. \\ 
- The question is delimited by <question></question>.\\
- A translation of the question in layman terms is delimited by <translated\_question></translated\_question>.\\
- Your answer should be delimited by <thought\_process></thought\_process>.\\
\\
\#\# Question \\
<question> \{question\} </question> \\
<translated\_question> \{translated\_question\} </translated\_question> \\

\#\# Thought process \\ 
}{Prompt to determine relevancy of reference child chunks}

\prompt{Filter -- Validation Prompt}{
\# Instruction \\
\\
\#\# Reference \\
<reference> \{context\} </reference> \\ 
\\
\#\# Question \\ 
<question> \{question\} </question> \\
<translated\_question> \{translated\_question\} </translated\_question> \\
\\
\#\# Thought Process\\
\{cot\_info\}\\
\\
\#\# Rules \\
- There are multiple reference texts.\\
- For each reference text, your task is to use the given thought process to decide whether the reference text can be used to answer the given question. \\
- If you need to cite the reference text to answer the question, reply with True \\
- If not, reply with False.\\
- Only include True and False per reference text\\
- Each reference text is delimited by <reference></reference>.  \\
- The question is delimited by <question></question>. \\
- A translation of the QUESTION in layman terms is provided within <translated\_question></translated\_question>.\\
- The thought process is delimited by <thought\_process></thought\_process>.
- Your answer is delimited <validation></validation>.\\
\\
\#\# Validation\\
}{Prompt to decide which child chunks to filter out}

\prompt{Basic Reader Prompt}{
\# Instruction \\
\hspace*{2em}You are an expert in corporate tax policies. \\

\#\# Rules \\
\hspace*{2em}- Generate an ANSWER to the provided QUESTION using the CONTEXT given to you. \\
\hspace*{2em}- While generating the ANSWER use the style of the examples provided below. \\
\hspace*{2em}- Ensure the ANSWER is incisive and concise. \\
\hspace*{2em}- Enclose your ANSWER within \textless answer\textgreater\textless /answer\textgreater \\

\#\# Context \\
\hspace*{2em}\{context\} \\

\#\# Chat History \\
\hspace*{2em}\{chat\_history\} \\

\#\# Example Questions and Answers \\
\hspace*{2em}\{examples\} \\

\#\# Question in Tax/Legal Lingo \\
\hspace*{2em}\{question\} \\

\#\# Question Translation in Layman Terms \\
\hspace*{2em}\{translated\_questions\} \\

\#\# Answer
}{Prompt for generating answers from the provided context}

\prompt{Domain-Specific Reader Prompt}{
\# Instruction \\
You are an expert in understanding documents containing tax opinions and advices. \\
    
\#\# Key Legal Phrases\\
<domain-specific data>\\
    
\#\# Rules\\
- Generate an ANSWER to the provided QUESTION using the CONTEXT given to you.\\
- While generating the ANSWER \\
\hspace*{2em}- If the QUESTION is about Tax Opinions, use Tax Opinion Related Phrases specified above\\
\hspace*{2em}- Additionally, use the style of the examples provided below. \\
\hspace*{2em}- The ANSWER will be read by tax lawyers, ensure it follows tax legal language and is incisive and concise.\\
- Enclose your ANSWER within <answer></answer>\\

\#\# Context\\
\{context\}\\

\#\# Chat History\\
\{chat\_history\}\\
    
\#\# Example Questions and Answers\\
\{examples\}\\

\#\# Question in Tax/Legal Lingo\\
\{question\}\\
    
\#\# Question Translation in Layman Terms\\
\{translated\_questions\}\\

\#\# Answer"""
}{Prompt for generating answers from the provided context}

\prompt{Evidence-based Question Generator}{
\# Instructions \\

\#\# Task \\
\hspace*{2em}- Your task is to generate evidence-based questions and also generate answers to the questions \\
\hspace*{2em}- Evidence-based questions help the user to learn about the features/description/definition of a concept/idea/object/event. \\
\hspace*{2em}- Answers to evidence-based questions should include wikipedia-like passage describing/defining an event/object or its properties based only on facts. \\

\#\# Rules \\
\hspace*{2em}- Generate 2 evidence-based questions by using a combination of facts from each chunk \\
\hspace*{2em}- For each question, you must generate the corresponding answer to the question using the chunks given \\
\hspace*{2em}- Each chunk is enclosed in \textless chunk\textgreater\textless /chunk\textgreater \\
\hspace*{2em}- First question should have NO named entities \\
\hspace*{2em}- Second question can have named entities \\
\hspace*{2em}- Both the questions should NOT be verbose \\
\hspace*{2em}- Examples of comparison questions are as follows: \\
\hspace*{4em}- What is ...? \\
\hspace*{4em}- How does/do ... work? \\
\hspace*{4em}- What are the properties of ...? \\
\hspace*{4em}- What is the meaning of ...? \\
\hspace*{4em}- How do you describe ...? \\
\hspace*{2em}- The output should be formatted as bullet points: \\
\hspace*{4em}- Question 1: \textless question1\textgreater \\
\hspace*{4em}- Answer 1: \textless answer1\textgreater \\
\hspace*{4em}- Question 2: \textless question2\textgreater \\
\hspace*{4em}- Answer 2: \textless answer2\textgreater \\

\#\# Text \\
\hspace*{2em}\{text\}
}{Prompt to generate evidence-based question and answer pairs}

\prompt{Comparison-based Question Generator}{
\# Instructions \\

\#\# Task \\
\hspace*{2em}- Your task is to generate comparison-based questions and also generate answers to the questions \\
\hspace*{2em}- Comparison based questions help the user to compare/contrast two or more things, understand their differences/similarities \\
\hspace*{2em}- Answers to comparison-based questions should include wikipedia-like passage describing/defining an event/object or its properties based only on facts. \\

\#\# Rules \\
\hspace*{2em}- Generate 2 comparison-based questions by using a combination of facts from each chunk. \\
\hspace*{2em}- For each question, you must generate the corresponding answer to the question using the chunks given \\
\hspace*{2em}- Each chunk is enclosed in \textless chunk\textgreater\textless /chunk\textgreater \\
\hspace*{2em}- First question should have NO named entities \\
\hspace*{2em}- Second question can have named entities \\
\hspace*{2em}- Both the questions should NOT be verbose \\
\hspace*{2em}- Examples of comparison questions are as follows: \\
\hspace*{4em}- How is X ... to/from Y? \\
\hspace*{4em}- What are the ... of X over Y? \\
\hspace*{4em}- How does X ... against Y? \\
\hspace*{2em}- The output should be formatted as bullet points: \\
\hspace*{4em}- Question 1: \textless question1\textgreater \\
\hspace*{4em}- Answer 1: \textless answer1\textgreater \\
\hspace*{4em}- Question 2: \textless question2\textgreater \\
\hspace*{4em}- Answer 2: \textless answer2\textgreater \\

\#\# Text \\
\hspace*{2em}\{text\}
}{Prompt to generate comparison-based question and answer pairs}

\prompt{Domain-specific Question Generator}{
\# Instructions \\

\#\# Task \\
\hspace*{2em}- As a corporate tax law expert, your task is to generate domain specific questions that use tax law language and also generate answers to the questions \\
\hspace*{2em}- Domain specific questions incorporate tax law terminology to ask questions that are of interest to domain experts \\
\hspace*{2em}- Answers to questions using tax law language should answer the question using facts found in the chunks \\

\#\# Rules \\
\hspace*{2em}- Generate 2 domain specific questions using tax law terminology by using a combination of facts from each chunk. \\
\hspace*{2em}- For each question, you must generate the corresponding answer to the question using the chunks given \\
\hspace*{2em}- Each chunk is enclosed in \textless chunk\textgreater\textless /chunk\textgreater \\
\hspace*{2em}- First question should have NO named entities \\
\hspace*{2em}- Second question can have named entities \\
\hspace*{2em}- Both the questions should NOT be verbose \\
\hspace*{2em}- Examples of tax law terminology include: \\
\hspace*{4em}- Level of comfort \\
\hspace*{4em}- Tax characterization \\
\hspace*{4em}- Witholding tax \\
\hspace*{2em}- The output should be formatted as bullet points: \\
\hspace*{4em}- Question 1: \textless question1\textgreater \\
\hspace*{4em}- Answer 1: \textless answer1\textgreater \\
\hspace*{4em}- Question 2: \textless question2\textgreater \\
\hspace*{4em}- Answer 2: \textless answer2\textgreater \\

\#\# Text \\
\hspace*{2em}\{text\}
}{Prompt to generate domain-specific question and answer pairs}

\prompt{Summarization for Question Generation}{
\# Instruction \\

\#\# Task \\
\hspace*{2em}- As a professional summarizer, create a concise and comprehensive summary of the provided text while adhering to these guidelines: \\
\hspace*{4em}- Craft a summary that is detailed, thorough, in-depth, and complex, while maintaining clarity and conciseness. \\
\hspace*{4em}- Incorporate main ideas and essential information, eliminating extraneous language and focusing on critical aspects. \\
\hspace*{4em}- Rely strictly on the provided text, without including external information. \\

\#\# Text \\
\hspace*{2em}\{text\}
}{Prompt to generate summaries for question generation}

\prompt{Instruction-based Question Generator}{
\# Instructions \\

\#\# Task \\
\hspace*{2em}- Your task is to generate instruction-based questions and also generate answers to the questions \\
\hspace*{2em}- Instruction-based questions help the user to understand the procedure/method of doing/achieving something \\
\hspace*{2em}- Answers to instruction-based questions involve instructions/guidelines in a step-by-step manner \\

\#\# Guidelines \\
\hspace*{2em}- Generate 2 different instruction-based questions by using a combination of facts from at least two chunks \\
\hspace*{2em}- For each question, you MUST generate the corresponding answer to the question using the chunks given \\
\hspace*{2em}- Each chunk is enclosed in \textless chunk\textgreater\textless /chunk\textgreater \\
\hspace*{2em}- First question should have NO named entities \\
\hspace*{2em}- Second question can have named entities \\
\hspace*{2em}- Both the questions should NOT be verbose \\
\hspace*{2em}- Examples of instruction-based questions are as follows: \\
\hspace*{4em}- How to ...? \\
\hspace*{4em}- How can we do ...? \\
\hspace*{4em}- What is the process for ...? \\
\hspace*{4em}- What is the best way to ...? \\
\hspace*{2em}- The output should be formatted as bullet points: \\
\hspace*{4em}- Question 1: \textless question1\textgreater \\
\hspace*{4em}- Answer 1: \textless answer1\textgreater \\
\hspace*{4em}- Question 2: \textless question2\textgreater \\
\hspace*{4em}- Answer 2: \textless answer2\textgreater \\

\#\# Text \\
\hspace*{2em}\{text\}
}{Prompt to generate instruction-based question and answer pairs}

\prompt{Reason-based Question Generator}{
\# Instructions \\

\#\# Task \\
\hspace*{2em}- Your task is to generate reason-based questions and also generate answers to the questions \\
\hspace*{2em}- Reason-based questions help the user to find out reasons of/for something \\
\hspace*{2em}- Answers to reason-based questions involve a list of reasons with evidence \\

\#\# Rules \\
\hspace*{2em}- Generate 2 reason-based questions by using a combination of facts from at least two chunks \\
\hspace*{2em}- For each question, you MUST generate the corresponding answer to the question using the chunks given \\
\hspace*{2em}- Each chunk is enclosed in \textless chunk\textgreater\textless /chunk\textgreater \\
\hspace*{2em}- First question should have NO named entities \\
\hspace*{2em}- Second question can have named entities \\
\hspace*{2em}- Both the questions should NOT be verbose \\
\hspace*{2em}- Examples of reason-based questions are as follows: \\
\hspace*{4em}- Why does ...? \\
\hspace*{4em}- What is the reason for ...? \\
\hspace*{4em}- What causes ...? \\
\hspace*{4em}- How come ... happened? \\
\hspace*{4em}- How can ...? \\
\hspace*{2em}- The output should be formatted as bullet points: \\
\hspace*{4em}- Question 1: \textless question1\textgreater \\
\hspace*{4em}- Answer 1: \textless answer1\textgreater \\
\hspace*{4em}- Question 2: \textless question2\textgreater \\
\hspace*{4em}- Answer 2: \textless answer2\textgreater \\

\#\# Text \\
\hspace*{2em}\{text\}
}{Prompt to generate reason-based question and answer pairs}

\prompt{List-based Question Generator}{
\# Instructions \\

\#\# Task \\
\hspace*{2em}- Your task is to generate list-based questions and also generate answers to the questions \\
\hspace*{2em}- List based questions help the user learn about the properties/requirements/components of a concept/law/idea \\
\hspace*{2em}- Answers to list-based questions should include a list that outlines the properties/requirements/componets defining a concept/law/idea, which is based on the facts found in the chunks. \\

\#\# Rules \\
\hspace*{2em}- Generate 2 list-based questions by using a combination of facts from each chunk. \\
\hspace*{2em}- For each question, you must generate the corresponding answer to the question using the chunks given \\
\hspace*{2em}- Each chunk is enclosed in \textless chunk\textgreater\textless /chunk\textgreater \\
\hspace*{2em}- First question should have NO named entities \\
\hspace*{2em}- Second question can have named entities \\
\hspace*{2em}- Both the questions should NOT be verbose \\
\hspace*{2em}- Examples of comparison questions are as follows: \\
\hspace*{4em}- What are the requirements of X? \\
\hspace*{4em}- What are the X of Y? \\
\hspace*{2em}- The output should be formatted as bullet points: \\
\hspace*{4em}- Question 1: \textless question1\textgreater \\
\hspace*{4em}- Answer 1: \textless answer1\textgreater \\
\hspace*{4em}- Question 2: \textless question2\textgreater \\
\hspace*{4em}- Answer 2: \textless answer2\textgreater \\

\#\# Text \\
\hspace*{2em}\{text\}
}{Prompt to generate list-based question and answer pairs}

\prompt{Closed Source Query Rewriter -- query to doc}{
Passage construction:
You are Corporate Tax expert. Given a query that is posed by a fellow corporate tax expert, your job is to write a passage that answers the given query.
  
-------------------------------------------------
 
EXAMPLES:
 
  EXAMPLE 1:
  Query: <domain-specific query>
  Passage: <domain-specific passage>
 
  EXAMPLE 2:
  Query: <domain-specific query>
  Passage: <domain-specific passage>
 
-------------------------------------------------
  
Query: <query>
Passage:
}{Closed-source query re-writer prompt to retrieve doc from query}

\prompt{Closed Source Query Rewriter -- doc to rewrite}{
Query generation:
You are an information retrieval expert. Given a query and a passage, which is relevant to the query, rewrite the query in 3 different ways to make it more clear and more descriptive.
--------------------------------------------------------
GOAL:
The goal is to help improve the retrieval of the right pieces of information that will help answer the query. By generating multiple perspectives of the rewritten query, your goal is to help
the user overcome some of the limitations of the distance-based similarity search.
    
---------------------------------------------------------
RULES AND GUIDELINES:
- Rewrite the query in 3 different ways
- Make sure that each rewritten query is only one single question. Do not expand it into multiple questions.
- The rewritten queries must add more context to the original query, using the passage given.
- Provide these alternative questions separated by newlines.

query: {query}
passage: {passage}
rewritten queries:
}{Closed-source query re-writer prompt for getting rewrite from generated passage}


\prompt{Modified claim extraction prompt}{
\# Instructions \\
- Given a question and a candidate answer to the question, please extract a KG from the candidate answer conditioned on the question. \\
- Represent the KG with triples formatted as ("subject", "predicate", "object"), with each triple in a single line. \\
- Please note that this is an EXTRACTION task, so DO NOT care about whether the content of the candidate answer is factual or not, just extract the triples from it. \\
- Importantly, ensure that the extracted KG does not contain overlapping or redundant information. Each piece of information should be represented in the KG only once, and you should avoid creating triples that are simply the inverse of another triple. \\

\# Clarification on Redundancy \\
- First, do not create triples that reverse the subject and object to state the same fact. \\
- Next, ensure each fact is represented uniquely in the simplest form, and avoid creating multiple triples that convey the same information. \\
- The facts can be of different levels of granularity such that it covers all of the knowledge in the answer. \\
- Do NOT break down the facts into lower granularity if it changes the semantics of the answer. \\
- Instead, the triples can be nested such that the subject and object can be triples themselves. \\

\# Examples \\
\hspace*{2em}- Question: Given these paragraphs about the Tesla bot, what is its alias? \\
\hspace*{2em}- Candidate Answer: Optimus (or Tesla Bot) is a robotic humanoid under development by Tesla, Inc. It was announced at the company's Artificial Intelligence (AI) Day event on August 19, 2021. \\
\hspace*{2em}- KG: \\
\hspace*{4em}- ("Optimus", "is", "robotic humanoid") \\
\hspace*{4em}- ("Optimus", "under development by", "Tesla, Inc.") \\
\hspace*{4em}- ("Optimus", "also known as", "Tesla Bot") \\
\hspace*{4em}- ("Tesla, Inc.", "announced", "Optimus") \\
\hspace*{4em}- ("Announcement of Optimus", "occurred at", "Artificial Intelligence (AI) Day event") \\
\hspace*{4em}- ("Artificial Intelligence (AI) Day event", "held on", "August 19, 2021") \\
\hspace*{4em}- ("Artificial Intelligence (AI) Day event", "organized by", "Tesla, Inc.") \\

\hspace*{2em}- Question: here is some text about Andre Weiss, how many years was Andre at University of Dijon in Paris? \\
\hspace*{2em}- Candidate Answer: 11 years \\
\hspace*{2em}- KG: \\
\hspace*{4em}- ("Andre Weiss at University of Dijon in Paris", "duration", "11 years") \\

\hspace*{2em}- Question: Are all corporate entities required to file quarterly tax returns under the new tax regulations? \\
\hspace*{2em}- Candidate Answer: No. While the new tax regulations generally require corporate entities to file quarterly tax returns, there are exceptions for small businesses and for non-profit organizations that meet specific criteria. \\
\hspace*{2em}- KG: \\
\hspace*{4em}- ("New tax regulations", "require", "corporate entities to file quarterly tax returns") \\
\hspace*{4em}- ("Exceptions", "exist for", "small businesses that meet specific criteria") \\
\hspace*{4em}- ("Exceptions", "exist for", "non-profit organizations that meet specific criteria") 

}{Prompt for extracting nested triples from a candidate answer}


\prompt{Correctness Evaluation using LLM-as-a-judge}{
\# Instructions \\

\#\# Task \\
\hspace*{2em}- You are an impartial judge \\
\hspace*{2em}- Given a QUESTION, a SOURCE and an ANSWER, your job is to evaluate whether the ANSWER answers the QUESTION correctly and completely. \\

\#\# Task Description: \\
\hspace*{2em}- Given a QUESTION, a SOURCE and an ANSWER, \\
\hspace*{4em}- You need to determine if the ANSWER answers the QUESTION correctly and completely. \\
\hspace*{4em}- The ANSWER answers the question correctly and completely if the claims present in the ANSWER are sufficient to answer the QUESTION. \\
\hspace*{4em}- Refer the SOURCE to determine the set of necessary claims required to answer the QUESTION. \\
\hspace*{4em}- To determine if the ANSWER correctly and completely answers the QUESTION \\
\hspace*{6em}- Extract a set of claims from the SOURCE which are necessary to answer the QUESTION \\
\hspace*{6em}- Determine if each of the necessary claims is present in the ANSWER \\
\hspace*{6em}- Generate your thought process to perform this task and enclose it in \textless thought\_process\textgreater\textless /thought\_process\textgreater \\
\hspace*{4em}- Based on your thought process, provide the final decision about whether the ANSWER answers the QUESTION correctly and completely \\
\hspace*{6em}- You would return INCORRECT if a claim in the ANSWER is incorrectly answering the QUESTION \\
\hspace*{6em}- You would return PARTIAL if a necessary claim is missing in the ANSWER \\
\hspace*{6em}- You would return COMPLETE if all of the necessary claims are present in the ANSWER \\
\hspace*{6em}- Enclose your final decision in \textless decision\textgreater\textless /decision\textgreater \\

QUESTION: \\
\hspace*{2em}\{question\} \\

SOURCE: \\
\hspace*{2em}\{source\} \\

ANSWER: \\
\hspace*{2em}\{answer\} \\

RESPONSE:
}{Prompt to use LLM-as-a-judge to evaluate answer correctness}

\end{document}